\documentclass[letterpaper, 10 pt, conference]{ieeeconf}  % Comment this line out if you need a4paper

\IEEEoverridecommandlockouts                              % This command is only needed if 
\title{\LARGE \bf Learning Input Constrained Control Barrier Functions for Guaranteed Safety of Car-Like Robots}

\usepackage[T1]{fontenc}
\usepackage[utf8]{inputenc}
\usepackage{authblk}
\usepackage{float}

\usepackage{amsthm}
\usepackage{amssymb}
\usepackage{amsmath,color}
\usepackage{graphicx}
\usepackage{subcaption}
\usepackage{stmaryrd}
\usepackage{mathbbol}
\usepackage{mathtools}
\usepackage{relsize}
\usepackage{paralist}
\usepackage{bm}
\usepackage{cite,float,algorithmic}
\usepackage[dvipsnames]{xcolor}
\usepackage[normalem]{ulem} % Cross out text
\usepackage{soul}
\usepackage{url}
\usepackage{placeins}  % float barrier
\let\labelindent\relax
\usepackage{enumerate}
\usepackage[labelformat=simple]{subcaption}  % For referencing subfigures: e.g. 1(a)
\usepackage{algorithm}
\usepackage{psfrag}
\usepackage{shellesc}
\usepackage{hyperref}

\usepackage{tikz}
\usepackage{pgfplots}
\pgfplotsset{compat=newest}
\pgfplotsset{plot coordinates/math parser=false}
\usetikzlibrary{external}
\usepgfplotslibrary{fillbetween}
\tikzset{/tikz/external/optimize=false}
\usetikzlibrary{shapes,arrows}
\usetikzlibrary{calc}
\usetikzlibrary{backgrounds}
\usetikzlibrary{fit}

\newcommand{\R}{\mathbb{R}}

\newcommand{\C}{\mathcal{C}}
\newcommand{\X}{\mathcal{X}}
\newcommand{\Z}{\mathcal{Z}}
\newcommand{\D}{\mathcal{D}}
\newcommand{\U}{\mathcal{U}}
\newcommand{\K}{\mathcal{K}}
\DeclareMathOperator*{\argmin}{arg\,min}
\DeclareMathOperator*{\inter}{int}

\newtheorem{defi}{Definition}

\newtheorem{thm}{Theorem}

\newtheorem{lem}{Lemma}

\newtheorem*{thm*}{Theorem}

\newcommand\scalemath[2]{\scalebox{#1}{\mbox{\ensuremath{\displaystyle #2}}}}

\newenvironment{sizeddisplay}[1]
 {\par\nopagebreak#1\noindent\ignorespaces}
 {\nopagebreak\ignorespacesafterend}

\newlength\fheight %tikz pic height
\newlength\fwidth  % tikz pic weight

\author{Sven Br{\"u}ggemann$^{1}$, Dominic Nightingale$^{1}$, Jack Silberman$^{2}$, and Maur{\'i}cio de Oliveira$^{1}$% <-this % stops a space
\thanks{$^{1}$Sven Br{\"u}ggemann, Dominic Nightingale, and Maur{\'i}cio de Oliveira are with the Mechanical \&\ Aerospace Engineering Department, University of California, San Diego, CA 92093-0411, USA {\tt\small \{sbruegge,djnighti,mauricio\}@eng.ucsd.edu}}%
\thanks{$^{2}$Jack Silberman is with the Department of Electrical \& Computer Engineering, University of California, San Diego, CA 92093-0411, USA
        {\tt\small jacks@eng.ucsd.edu}}%
}
\begin{document}
\maketitle

%===============================================================================

\begin{abstract}
     We propose a design method for a robust safety filter based on Input Constrained Control Barrier Functions (ICCBF) for car-like robots moving in complex environments. A robust ICCBF that can be efficiently implemented is obtained by learning a smooth function of the environment using Support Vector Machine regression. The method takes into account steering constraints and is validated in simulation and a real experiment.
\end{abstract}

%===============================================================================

\section{Introduction}
	
	Safety is often prioritized over performance criteria due to physical limitations of the system leading to unsafe operation (e.g. actuator saturation, power constraints, etc), and/or safety requirements imposed as a result of interacting with the environment. Examples of the latter are obstacle avoidance \cite{aljalbout2021learning, pandey2017mobile}, lane-keeping for autonomous vehicles \cite{sallab2016end} and navigation tasks \cite{chou2020uncertainty}.
	
\begin{figure}[t]
        \centering
        \begin{subfigure}[t]{0.9\columnwidth}%{0.65\columnwidth}
            \centering
            \includegraphics[width=0.9\columnwidth]{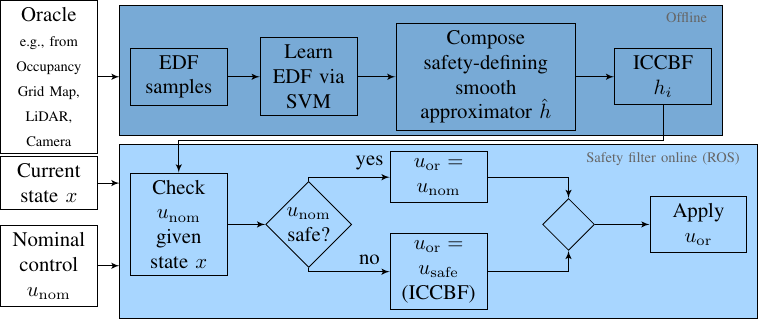}
            \caption{Proposed safety filter design process.\label{fig:diagram}}
        \end{subfigure}
        % \hfill
        % \begin{subfigure}[t]{0.7\columnwidth}%{0.33\columnwidth}
        %     \centering
        %     \includegraphics[width=.4\columnwidth]{figures/car_small.jpg}
        %     \includegraphics[width=.2\columnwidth]{figures/IMG_5747_small.jpg}
        %     % \includegraphics{figures/car.jpeg}
        %     \caption{1:10 RC car and track.
        %     \label{fig:car+track}}
        % \end{subfigure}
        % \hfill
        % \begin{subfigure}[t]{0.15\columnwidth}
        %     \centering
        %     \includegraphics[width=.7\columnwidth]{figures/IMG_5747.jpeg}
        %     \caption{Race track we wish to safely follow.
        %     \label{fig:exp_race_track}}
        % \end{subfigure}
        \caption{Overview of approach and photos of experiment.
            \label{fig:h_exp}}
    \end{figure}

	Recently, Control Barrier Functions (CBFs) \cite{ames:cbf_intro} have become a popular strategy to enforce safety.
 % They provide formal safety guarantees using the notion of forward invariance: if the (robot) state starts in the safe set, it will remain in the safe set if appropriate action (or control) is taken. %(we use the terms action and control interchangeably).
	% Hence, CBFs-based methods can be seen as the dual to Control Lyapunov Function (CLF)-based approaches for stability.
	%but, in contrast, allow the states to move outside sub-level sets [https://coogan.ece.gatech.edu/papers/pdf/amesecc19.pdf].
	%
	% Another advantage of CBFs is that they can be imposed as an additional constraint on top of other performance-oriented objectives. 
% 	Applications include navigation algorithms for motion planning and trajectory tracking tracking while guaranteeing safety by avoiding obstacles  \cite{dawson2021safe,li2021instantaneous,yang2019sampling}.
% 	   \item trajectory tracking algorithms minimize the distance to a given trajectory, while guaranteeing safety by avoiding dynamic obstacles \cite{dawson2021safe};
% 	    \item {safety filters} apply the nominal action or if unsafe a safe action that is as close as possible to the nominal case. The nominal action is for example provided by the trajectory planning/tracking algorithm or by a manual operator.
% 	\end{enumerate*}
	One common setup for CBFs is that of \emph{supervising} a given nominal action which may not necessarily be safe. The CBF acts as a \emph{safety filter} which computes an overriding safe action that is as ``close as possible'' to the nominal one if safety is at risk.
 % The nominal action may be provided by a trajectory planning or tracking algorithm, a manual operator, or a potentially unsafe neural network approach. 
	Yet, whereas CBFs provide a strong theoretical framework upon which to build safety-critical systems, there are still practical/implementation difficulties, and many approaches are validated in simulation only \cite{ctx28136798090006531,
    %choi2020reinforcement,
 9303785,9294485,lindemann2021learning,li2021instantaneous,ferlez2020shieldnn,barbosa-edf_gauss,Srinivasan2020SynthesisOC}. {One of these difficulties arise for systems of high relative degree which naturally arise for car-like robots as safety is commonly expressed in terms of $x$-$y$ coordinates only, excluding the steering angle.} This work is concerned with the practical deployment of a {supervising} Input Constrained CBF (ICCBF) based safety filter for car-like robots. The following aspects will be considered in detail:
	%\vspace{-.25cm}
	\begin{itemize}
	    \item [\textbf{CBF Learning}] directly from {the Euclidean Distance Function (EDF) encoding} a global map in the form of a kernelized Support Vector Machine (SVM) representation (loosely speaking, EDFs represent the shortest distance to any unsafe point in space); 
	    %\vspace{-0.1cm}
	    \item [\textbf{Robustification}] { in practice against uncertainties including noisy data, measurement and process noise, discretization and uncertain dynamics (see Section \ref{sec:sim_exp})} to ensure safe operation based on the resulting smooth SVM approximation;
	    %\vspace{-0.1cm}
	    \item [\textbf{Input Constraints}] resulting from robot steering limitations are handled by {a reformulated model and} an ICCBF safety filter;
	    %\vspace{-0.1cm}
	    \item [\textbf{Validation}] via simulation;
	    %\vspace{-0.1cm}
	    \item [\textbf{Implementation}] on a 1:10 RC car.
	\end{itemize}
	A diagram of the blocks involved in the complete process is given in Fig.~\ref{fig:diagram}. {Notice that though we explicitly learn the EDF via SVM, we implicitly also learn the related ICCBF, as gradients are available in closed form. While we choose to learn the EDF, our method is also applicable to other map representations such as \cite{Oleynikova2016SignedDF}}. 
	%As a result of the proposed approach, a car-like robot will have its operation overridden by the CBF-based safety filter if the given nominal steering action is deemed unsafe, which here means that the robot collides with an obstacle or goes off road.
	
% 	An Euclidean Distance Function (EDF) will be learned from a global map and robustifyed with the introduction of a robustness parameter (loosely speaking EDFs represent the shortest distance to any unsafe point in space). A CBF constructed using the EDF is then made resilient to the robot's steering limitations
	
% 	The resulting CBF does not take into account the vehicle's steering limitations. 
% 	We apply our approach in simulation and real experiments using a 1:10 RC car. The procedure is illustrated in Figure \ref{fig:diagram}: offline, from some oracle we obtain EDF samples which are used to learn a smooth approximator. The approximator is then used to compose the safe set-defining output function. This is then robustified and adapted to the input constrained case to alleviate the issue of infeasibility. Finally, the learned safety filter is used online to interrupt nominal operation if safety cannot guaranteed otherwise, using the current robot state.
	% \setlength{\belowcaptionskip}{-10pt}
    % \setlength{\belowcaptionskip}{-3pt}
    %\vspace{-.3cm}
    % \begin{figure}[t]
    %     \centering
    %     \includegraphics[trim=0 130 0 130,clip,width=.65\columnwidth]{figures/diagram_vert5.pdf}
    %     \caption{Overview of proposed learned safety design.\label{fig:diagram}}
    % \end{figure}
    
    % \begin{figure}[t]
    %     \centering
    %     %\resizebox{\columnwidth}{!}{
    %     \includegraphics[width=.5\columnwidth]{figures/diagram_vert2.pdf}%}
    %     % %\vspace{-1.5cm}
    %     \caption{Overview of proposed learned safety design. From some oracle we obtain EDF samples $\{\mathcal{D}_i\}_{i=1}^{n_s}$, which are used to learn (via SVM) the CBF offline which encodes safety for the environment. The learned CBF is then used online to interrupt nominal operation if safety cannot guaranteed otherwise, and applies a control that renders the system safe.
    %     \label{fig:diagram}}
    % \end{figure}

	\paragraph*{Contribution} 
	ICCBF-based filter design depends on the explicit knowledge of a valid ICCBF. This is challenging in practice, especially for car-like robots with actuator constraints moving in complex unsafe regions; complex here means areas that cannot be described by simple geometries like ellipsoids. We overcome this difficulty by learning a robust, smooth approximator of the EDF {(whose explicit form is unknown)} with safety guarantees for arbitrary geometries while considering steering constraints. {Unlike most works on CBFs, we consider high-relative degree systems, and towards} implementation we develop a closed-form, vectorized algorithm which outperforms a naive formulation by three orders of magnitude {(see computing time in Section~\ref{sec:sim_exp})}. {Essential for the implementation of the algorithm is our reformulation of the vehicle model which reduces the number of input constraints to one.} We present a real-world experiment to provide evidence that simulations are transferable to real robots. %Both the vectorized algorithm and the ROS package developed for the experiment are available in \cite{sven_code_github}.
    {The comprehensive approach from learning from data all the way to the derivation of an implementable, experimentally validated algorithm with much attention to detail is our main contribution.}

    \paragraph*{Related work}
    Learning and CBF-based design usually assumes that CBFs are known but other ingredients such as the system dynamics are uncertain (see e.g. \cite{taylor:learning_CBF,ctx28136798090006531}%choi2020reinforcement
    ). Learning the CBF itself was proposed in \cite{9303785,Saveriano:learning_CBF}, where expert demonstrations are used to derive safety guarantees. \cite{jin-neural_certificates} avoids the assumption of control-affine dynamics by jointly learning the Control Lyapunov Function (CLF), CBF and related policy. Yaghoubi et al. \cite{9294485} propose imitation learning to learn the CBF-based controller, avoiding solving the related QP. In \cite{lindemann2021learning} data is used to learn certifiable safe control laws based on CBFs for hybrid systems. Removing the assumption of known dynamics for multi-agent systems is the focus of \cite{ctx28136798090006531}. Learning decentralized CBFs for the safety of multi-agent systems is also treated in \cite{qin-multiagent}. A model-based learning approach to synthesize robust feedback control was presented in \cite{dawson2021safe}. The authors of \cite{li2021instantaneous} learn the CBFs from sensor data and combine it with the CLFs as part of a QP. A recent overview of the rapidly growing field of learned safety certificates is \cite{dawson2022safe}.
    
    Most related to our work are \cite{ferlez2020shieldnn,barbosa-edf_gauss,Srinivasan2020SynthesisOC,9392327}. The authors of \cite{ferlez2020shieldnn} develop a neural network (NN) safety filter for bicycle models facing stationary, fixed-radius obstacles in the plane. {Tensor gradients of the NN required for implementation tend to be more expansive than our vectorized formulation. For obstacle avoidance, in \cite{9392327} NNs are trained online to approximate the distance function, based on which CBFs are directly formulated as a convex second-order cone problem, which is usually more costly than an explicit QP solution.}
    Learning the EDF (and its derivatives) for CBF design was proposed in \cite{barbosa-edf_gauss}. Therein, Gaussian Process was used as a supervised learning strategy requiring derivative measurements (or corresponding numerical approximations).
    % Theoretical results assume that the control-affine term in the system dynamics is full rank. 
    % Computation involves matrix inversion with time complexity of $O(n_s^3)$, with $n_s$ being the number of samples.
    %Results are validated via simulations in Matlab.  
    Towards safety for obstacle avoidance \cite{Srinivasan2020SynthesisOC} uses SVM classification to learn the CBF based on sensor data (on- and offline). However, derivatives of the classifier required for the CBF design {may vanish or be unbounded with related numerical instabilities or loss of safety (see Section \ref{sec:learning}). In contrast to the present paper, none of these {learning-based} methods directly consider systems of high relative degree with input constraints; nor do they discuss implementation details or present real-world validation and related code.}
    
    % Further, computational aspects and experimental validation are absence and simulations do not reveal when the nominal control is overridden to ensure safety.

\section{Setup}\label{sec:background}
    %In this section we introduce the notion of safety and discuss the motivation for ICCBFs for constrained systems of higher relative degree, such as car-like robots. We then present the safety filter in closed form. 
    Consider the control-affine system,
    \vspace{-.15cm}
    \begin{align}\label{eq:sys_generic}
        \scalemath{1}{\dot x=f(x)+g(x)u},
    \end{align}
    with $x\in\X\subset\R^{n_x}$ as the current state and control input $u\in\U$ for some compact constraint set $\U\subset \R^{n_u}$, and smooth functions $f:\X\to\R^{n_x}$ and $g:\X\to\R^{n_x}\times\R^{n_u}$. 
    Let safety be described by a \emph{safe set} defined as the 0-superlevel set of a {sufficiently-often differentiable} function~$h(x):\R^{n_x}\to \R$:
    \vspace{-.15cm}
    \begin{align}\label{eq:safety_set}
        \C\doteq \{x\in\R^{n_x}:h(x)\geq 0\},
    \end{align}
    and write $\partial\C\doteq  \{x\in\R^{n_x}:h( x)= 0\},\, \inter(\C)\doteq  \{x\in\R^{n_x}:h( x)> 0\}.$ We assume that $\C$ is non-empty, closed, and simply connected.
    %System \eqref{eq:sys} with output function $h(\bar x$) is of relative degree two, i.e., $L_gL_fh(\bar x)=0$, with $L_fh$ denoting the Lie derivative of $h$ along the vector filed $f$, and $L_g$ accordingly. Note that $h(\bar x)$ only depends on $\bar x$ but its Lie derivatives depend on the entire state $x$.
    \begin{defi}[Forward Invariance \& Safety]\label{def:safe}
        The set $\C$ is forward invariant if $x_0\doteq x(t_0)\in \C$ at time $t_0$ implies that $x(t)\in\C$ for all time $t\geq t_0$. The system is considered safe with respect to $\C$ if $\C$ is forward invariant. 
    \end{defi}
    % The definition above applies to continuous-time systems. As we will implement our algorithm on the vehicle, we also introduce the discrete-time equivalent. 
    % \begin{defi}[DT Forward Invariance \& DT Safety]\label{def:safe}
    %     The set $\C$ is discrete-time forward invariant if $x_0\doteq x(t_0)\in \C$ at time $t_0$ implies that $x(t_k)\in\C$ for all time $k\geq 0,k\in\N$. The system is considered discrete-time safe if $\C$ is forward invariant. 
    % \end{defi}
    
    % Our CBF design uses the concept of class-$\K$ functions and that of relative degree which we both briefly present next. Class-$\K$ functions characterize scalar continuous functions, $\alpha(l)$, defined for $l\in[0, m)$ for some $m>0$ which are strictly increasing and $\alpha(0) = 0$. On the other hand, 
    
    % \begin{defi}[Relative degree] A system \eqref{eq:sys_generic} with output function $h(x)$ is of relative degree $r\in\N$ at point $x_o$ if for all $x$ in the neighborhood of $x_o$,
    % \begin{align}\label{eq:rel_deg}
    %     L_gL_f^kh(x) &=0, \quad k\leq r-2, &
    %     L_gL_f^kh(x) &\neq 0, \quad k\geq r-1
    % \end{align}
    % where $L_f$ denote the Lie derivative of $h(x)$ along the vector {field} $f$, and $L_g$ accordingly.
    % \end{defi}

    For common vehicle models, safety is often specified in terms of the vehicle position only, without considering heading information~\cite{8114339,covorsi,sven_drew_cbf}. This means that CBF-based safety design for car-like robots must overcome two hurdles: it needs to deal with a relative degree larger than one \emph{and} vehicle input constraints, which are not directly expressed in terms of the robot's position, must be enforced (e.g. steering limitations), potentially compromising feasibility.
    Since CBF safety is often formulated for relative-degree-one systems (e.g. \cite{ames:cbf_intro,8405547,ames2016control,ROMDLONY201639}), 
    % $h(x)$ as a CBF if there exists an extended class $\mathcal{K_{\infty}}$ function such that forward invariance is guaranteed \cite{ames:cbf_intro}. 
    % However, safety for land vehicles is usually expressed only in terms of the position rather than their entire state. 
    % Additionally, vehicle actuators such as servos and motor with limited torque impose control input constraints; this may lead to {an invalid CBF, i.e. to} infeasibility of the CBF safety design. 
    here we recall results on ICCBF from \cite{agrawal2021safe} which cover constrained systems with higher relative degree.
    %
    % \subsection*{Input Constrained Control Barrier Function}
    Consider system \eqref{eq:sys_generic} with constrained input $u\in\U$ and safe set $\C$. Let
    \begin{sizeddisplay}{\small}
    \begin{align}\label{eq:hi}
        h_0(x)&\doteq h(x),\nonumber\\
        h_1(x)&\doteq \inf_{u\in\U} \{L_fh_{0}(x)+L_gh_{0}(x)u+\alpha_0(h_{0}(x))\}\\[-4mm]%\,i=\{1,\dots,N\},
        &\vdots\nonumber\\[-2mm]
        %\vspace{-.2cm}
        h_N(x)&\begin{multlined}[t]
            \doteq\inf_{u\in\U}\{ L_fh_{N-1}(x)+L_gh_{N-1}(x)u +\alpha_{N-1}(h_{N-1}(x))\},
        \end{multlined}\nonumber
    \end{align}
    \end{sizeddisplay}
    where each $\alpha_i$ is a class-$\K$ function (see \cite[Section 4, Def. 4.2]{Khalil:1173048}) and $N\geq r-1$, with $r$ as the relative degree of \eqref{eq:sys_generic} with output function $h(x)$. The case $N=r-1$ leads to a standard exponential CBF~\cite{nguyen2016exponential}. Define the related safety sets as
    \vspace{-3mm}
    \begin{align}
        % \C_0&=\{x\in\X:h_0\geq0\}=\C,\\
        \C_i&=\{x\in\X:h_i(x) \geq0\},
        % &\vdots\\[-2mm]
        % \C_N&=\{x\in\X:h_N \geq0\}.
    \end{align}
    with $\C_0=\C$ and assume that $\C^\star = \C_0\cap\C_1\cap \cdots\cap \C_N$
    is not empty, closed and simply connected.
    \begin{defi}[\cite{agrawal2021safe}]\label{def:ICCBF}
        For system \eqref{eq:sys_generic} with safe set $\C$ and continuously differentiable class-$\K$ functions $\alpha_0,\dots,\alpha_{N-1}$, if there exists a class-$\K$ $\alpha_N$ such that
        \begin{sizeddisplay}{\small}
        \begin{align}
            \sup_{u\in\U}\{L_fh_N(x)+L_gh_N(x)u+\alpha_N(h_N(x))\}\geq0 \;\forall x\in\C^\star,
        \end{align}
        \end{sizeddisplay}
        then $h_N(x)$ is an {\normalfont Input Constrained Control Barrier Function (ICCBF)}.
    \end{defi}
    
    Importantly, $\C^\star \subset\C$, so the idea of ICCBFs is to shrink the safe set with each $\C_i$ until safety is guaranteed in the presence of bounded control inputs, as for all $x\in\C^\star$ \emph{any} control for which $h_N(x)\geq 0$ also ensures $h_{N-1}(x)\geq 0$, and eventually $h_0(x)=h(x){\geq0}$. Note that the need to perform the minimization in \eqref{eq:hi} on each $h_i(x)$ can be very difficult. One of the contributions of this paper is to show how this can be effectively done for car-like robots with steering constraints; see the last paragraph of Section \ref{sec:learning}. 
    % In this way, although the existence of an upper bound for $N$ has not been proved, ICCBFs increase robustness as possibly $h_0(x)=h(x)\geq 0$ even if $h_N(x)\leq 0$.
    
    %  \subsection*{Safety filter design with input constraints}
    % The definition of ICCBF resembles that of (zeroing) CBFs, and enables similar control designs: 
    Provided a valid ICCBF, let the nominal closed-loop be driven by control input $u_{\rm nom}(x)$, which may not be safe. A {safety filter} with overriding control, $u_{\rm or}(x)$, interrupts nominal operation if safety cannot be guaranteed is obtained through:
    \vspace{-.3cm}
    \begin{sizeddisplay}{\small}
    \begin{subequations}
        \begin{align}\label{eq:QP}
        u_{\rm or}(x)&=\argmin_{u\in\U}\|u_{\rm nom}(x)-u\|^2,\\ &\text{s.t. } L_fh_N(x)+L_g h_N(x) u + \alpha_N(h_N(x))\geq0.
    \end{align}
    \end{subequations}
    \end{sizeddisplay}%
    {Note that vanishing $L_gh_N(x)$ (due to, e.g., a vanishing partial derivatives of $h_N(x)$) risks safety as the inequality constraint would be independent of control $u$.}
    It follows from {smoothness implying locally Lipschitz} and \cite[Theorem 1]{agrawal2021safe} that the above problem admits the closed-form solution.
    \begin{lem}\label{lem:ICCBF_control}
        Consider system \eqref{eq:sys_generic} with input constraints $\U$ and $u_{\rm nom}\in\U$. Define
        \vspace{-.15cm}
        \begin{align*}
            u_{\rm safe}(x)=-L_g h_{N}(x)^{-1}(L_fh_{N}(x)+  \alpha_N(h_{N}(x)).
        \end{align*}
        If $h_N(x)$ is an ICCBF and $L_g h_{N}(x)$ is full rank, then
        \begin{sizeddisplay}{\small}
        \begin{align}\label{eq:explicit_QP}
            u_{\rm or}(x)=\begin{cases} u_{\rm nom}(x), & \begin{multlined}[t]
                L_fh_{N}(x)+L_g h_{N}(x)u_{\rm nom} \\[-.3cm]
                +\alpha_N(h_{N}(x))\geq0,
            \end{multlined} \\
            u_{\rm safe}(x),
             &\text{otherwise,} \end{cases}
        \end{align} 
        \end{sizeddisplay}
        renders \eqref{eq:sys_generic} safe. If $L_g h_{N}(x)$ is not full rank but $L_fh_{N}(x)+ \alpha_N( h_{N}(x))\geq0$ then \eqref{eq:sys_generic} is still safe.
    \end{lem}
    {The proof is omitted due to space constraints.} We note that {if the rank condition on $L_gh_N(x)$ cannot be guaranteed we can numerically solve the QP in \eqref{eq:QP}.} For the car-like robot however, this reduces to $L_gh_N(x)\in\R$ not being zero. 
    %The above concepts of safety will be applied to car-like robots in the next sections.
    
    \section{Safety design for car-like robots}
    \label{sec:safety_design}
    
    Assume a car-like robot that moves in a confined planar space described by compact $\Z\subset \R^2$ and suppose certain positions in $x$- and $y$-coordinate within $\Z$ are free, denoted by the set $\Z_{\rm free}\subset\Z$, and others are occupied, written as $\Z_{\rm occupied}$. We wish to ensure safety by avoiding $\Z_{\rm occupied}$ using the ICCBF design from Lemma \ref{lem:ICCBF_control}. {Accordingly, $\Z_{\rm occupied}$ is associated with $h(x)<0$.}
    
    %\vspace{-1ex}
    \paragraph*{Characterization of safe regions via Euclidean Distance Function}
    
    Safety is characterized by avoidance of occupied map positions, which represent obstacles and off-road areas. These safe/unsafe regions can be related to the EDF $d: \Z \to \R$,
    % Suppose the nominal feedback control in form of a steering command is given and does not necessarily avoid unsafe regions. 
    % Further, assume that data of the EDF are available: denote the $n_s$ samples as $\D=\{\D_i\}_{i=1}^{n_s}$, with $\D_i=(\bar x_i, d(\bar x_i))$, where the EDF $d:\C\to\R_{\geq0}$ is defined as
    \vspace{-.25cm}
    \begin{align}\label{eq:edf}
        d(z)\doteq \inf_{z'\in \Z_{\rm occupied}}\|z'-z\|, % \times \begin{cases} 1 \quad &\text{if } \bar x\in\C,\\
        % -1 \quad &\text{otherwise.}
        % \end{cases}
    \end{align}
    which describes the shortest distance to any occupied point in $\Z$. The EDF cannot be directly used to construct an ICCBF because it is not smooth enough. Instead, a suitable approximation will be learned in Section~\ref{sec:learning}.
    %Next we present the robot model required by our safety filter.
    
    %\vspace{-1ex}
    \paragraph*{Kinematic model with steering state and input constraints}
    
    We assume a constant positive forward velocity, $v_f$, and model the behavior of the front wheels (rather than the rear wheels or center of gravity) using the kinematic bicycle model \cite[Section 2, Table 2.1]{rajamani2011vehicle}:
    \begin{sizeddisplay}{\small} 
    \begin{align} \label{eq:sys}
        \begin{pmatrix} \dot x_{\rm f} \\ \dot y_{\rm f} \\ \dot \theta \\ \dot \delta\end{pmatrix}=\begin{pmatrix} v_{\rm f}\cos(\theta+ \delta)\\
        v_{\rm f}\sin(\theta+ \delta)\\
        v_{\rm f}\sin(\delta)/L\\
        u
        \end{pmatrix},
    \end{align}
    \end{sizeddisplay}
    where $x_{\rm f}$ and $y_{\rm f}$ are the $x$- and $y$-position of the front axle, $\theta$ the robot heading, $\delta$ the steering angle and $\dot \delta$ the front steering rate. We choose the control input to be the control signal $u=\dot \delta$ in order to ensure a continuous steering action. Doing so also makes the system control-affine, which is required by the proposed safety filter. We assume that both the steering angle and the angular steering rate are bounded as in:
    \vspace{-.2cm}
    \begin{align}
        |\delta|\leq \delta_{\rm max},\quad |u|\leq u_{\rm max}.
    \end{align}
    % The constraint on the steering angle could directly be considered by, e.g., defining an ICCBF of $h(\delta)=\delta_{\rm max}^2-\delta^2$. It could then be combined with other safety criteria related to the vehicle position: the QP in \eqref{eq:QP} would have one inequality constraint for each criterion. However, the steering limit would not affect the safe control associated with other criteria \emph{until} saturation occurs; this would likely lead to infeasibility and loss of safety. Thus, 
    
    \paragraph*{Modified model with only input constraints}
     Position \emph{and} steering constraints pose difficulties for the framework of ICCBFs\footnote{It is possible to impose multiple safety-related constraints, at the expense of feasibility guarantees. See the control-sharing property~\cite{xu2018constrained}, which is often difficult to establish.}. It also complicates the required minimization in \eqref{eq:hi} (see discussion at the end of Section \ref{sec:learning}). For this reason, we reduce the number of constraints to one by parameterizing the steering angle through a change in coordinates using the modified sigmoid function:
    \vspace{-.3cm}
    \begin{align}
        \delta \doteq \phi(\zeta)\doteq\delta_{\rm max}\left(\frac{2}{1+e^{-\zeta}} -1\right).
    \end{align}
    We consider the modified model
    \begin{sizeddisplay}{\small} 
    \begin{align} \label{eq:sys_aug}
        x &= \begin{pmatrix} x_{\rm f} \\ y_{\rm f} \\ \theta \\ \zeta\end{pmatrix}, &
        \dot { x} = \begin{pmatrix} \dot x_{\rm f} \\ \dot y_{\rm f} \\ \dot \theta \\ \dot \zeta\end{pmatrix}&=\begin{pmatrix} v_{\rm f}\cos(\theta+ \phi(\zeta))\\
        v_{\rm f}\sin(\theta+\phi(\zeta))\\
        v_{\rm f}\sin(\phi(\zeta))/L\\
        \left(\frac{\partial \phi(\zeta)}{\partial \zeta}\right)^{-1} u
        \end{pmatrix},
    \end{align}
    \end{sizeddisplay}%
    which ensures that the bounds on the steering angle and steering rate will not be exceeded if the single input constraint, 
    \vspace{-.4cm}
    \begin{align}
        |u|\leq u_{\rm max},
    \end{align}
    is satisfied. The next section shows how we can learn an ICCBF for the above model.

\section{{Learning a }robust safety filter}
\label{sec:learning}
    Safety as defined above relies on the output function, $h(x)$, whose sign characterizes the safe/unsafe regions. 
    Even when $h(x)$ is known, it may be difficult to calculate a suitable ICCBF satisfying Definition \ref{def:ICCBF}. Challenges include the need for $h(x)$ to be available in closed form and to be \emph{continuously differentiable} enough to define the output function $h_N(x)$ when $N > 1$. The EDF in \eqref{eq:edf} accurately defines the safe/unsafe regions but is not continuously differentiable and usually not available in closed form. Yet, samples can be generated, for example, from LiDAR measurements through test runs, occupancy grid representations, and satellite images. Assuming availability of such data, in this section we propose to use an SVM as a smooth surrogate for the EDF. Through an additional \emph{robustness parameter} we also provide theoretical safety guarantees of the learned safety filter. Lastly, we reveal additional computational advantages of the safety filter which are key to practical implementation.
    %\vspace{-1ex}
    \paragraph*{SVM smooth approximation to the EDF}
    
    We learn the EDF via $\epsilon$-SVM regression and then use the data to \emph{robustify} the related output function. Reasons for using a regression model instead of SVM classification as used in \cite{Srinivasan2020SynthesisOC} include the gradients needed for the safety filter. A classification model has an infinite gradient at the barrier and zero gradients in the safe regions, which likely renders the ICCBF invalid due to rank deficiencies in the term $L_g h_N( x)$. In contrast, a regression model has smaller gradients at the boundary and generically non-zero gradients inside the safe region. {Further, the simplicity of the SVM model allows us to obtain an efficient expressions for the tensor gradients as part of the safety filter in \eqref{eq:explicit_QP} which enables implementation on resource-limited robot hardware.}
    
    The basic idea of $\epsilon$-SVM regression is to find a function that is at least $\epsilon$-close to the all training samples and as flat as possible. The samples themselves serve as function parameters (support vectors), only if they are outside the $\epsilon$-tube. Importantly, the dual problem only involves dot products of samples so that for nonlinear regression, dot products of nonlinear functions $\Phi(z)$ can be implicitly described by kernel functions $k(z,z')=\langle \Phi(z), \Phi(z')\rangle$. Consider training data $(d_i,z_i)_{i=1}^{n_s}$ related \eqref{eq:edf}. The primal is defined as follows:
    \vspace{-.5cm}
    \begin{sizeddisplay}{\small} 
    \begin{subequations}\label{eq:svm_primal}
        \begin{align}
            \min_{w,a,\xi^{*},\xi} \quad & \frac{1}{2} w^\top w +G\sum_{i=1}^{n_s}(\xi_i+\xi_i^*),\\[-.1cm]
            \text{s.t. } \quad & -\epsilon - \xi_i^* \leq d_i - w^\top\Phi(z_i)-a\leq \epsilon + \xi_i,\\
            & \xi_i^{*},\xi_i\geq 0,
            % & w^\top\Phi(z_i)+a -d_i(z) \leq \epsilon + \xi_i^*,\\
        \end{align}
    \end{subequations}
    \end{sizeddisplay}
    where $\xi_i^*$ and $\xi_i$ are slack variables. The parameter $G>0$ balances flatness in the feature space against how samples outside the $\epsilon$ region are tolerated. We use the standard radial basis kernel,
    \vspace{-.2cm}
    \begin{align}
        k(z,z')=e^{-\gamma \|z-z'\|^2},
    \end{align}
    with $\gamma>0$ describing the impact of each sample on its neighboring points during training. Solving \eqref{eq:svm_primal} via its dual generates a smooth {(hence locally Lipschitz)} approximator for the EDF:
    \vspace{-.2cm}
    \begin{align}
        \label{eq:dhat}
        \hat d(z)=\sum_{i=1}^{n_{\rm sv}}\kappa_i k(z_i,z)+a,
    \end{align}
    with $n_{\rm sv}\leq n_{s}$ as the number of support vectors and $\kappa\in\R^{n_{\rm sv}}$ dual coefficients. Typically, the number of support vectors reduces as $\epsilon$ increases and $C$ decreases \cite{smola2004tutorial}.
    %\vspace{-1ex}
    \paragraph*{Robust output function and the ICCBF}
    
    Having obtained the smooth approximation $\hat{d}$, we define the output function as
    \begin{align}\label{eq:hat_h}
        h(x) &= \hat h(x) - \beta, & 
        \hat h(x) &= \hat d(C x), &
        C &= \begin{bmatrix} I_2 & 0 \end{bmatrix},
    \end{align}
    in which $I_2$ denotes a $2 \times 2$ identity matrix and $\beta\geq 0$ is a {robustness parameter}. Constructing $h_i(x)$, $\hat{h}_i(x)$, $i=0,\dots,N$, as in \eqref{eq:hi}, it follows that $h_i = \hat h_i - \alpha_i\circ\alpha_{i-1}\circ\dots\circ \alpha_0\circ \beta$,
    % \begin{align}\label{eq:h_robust_comparison}
    %     h_i = \hat h_i - \alpha_i\circ\alpha_{i-1}\circ\dots\circ \alpha_0\circ \beta, \quad i= 1,\dots,N,
    % \end{align}
    {where is $\circ$ the function composition operator with $\alpha_i\circ \alpha_{i-1}\doteq \alpha_i (\alpha_{i-1})$.}
    In {\cite{8405547} it is shown that a change in the level set, akin to that induced by $\beta$, provides robustness against input disturbances}; see also {Section \ref{sec:sim_exp} for other sources of uncertainties. Appropriate selection of $\beta$ and $\epsilon$ involves data and uncertainty analysis. Note that as common, $h_0(x)$ characterizes safety only in terms of the $x$-$y$ coordinates, excluding heading and steering information, whereas $h_r$ is the first barrier which explicitly considers steering rate constraints, alleviating the burden of infeasibility of the QP.} 
    
    The following theorem states that if the {approximation of the EDF is sufficiently accurate}, then the robust safety filter with control law \eqref{eq:explicit_QP} ensures safety for the robot.
    % Further examples requiring robustness are inaccurate approximations due to noisy data, errors in the modeled dynamics, and tolerated training samples inside the $\epsilon$-tube (set $\beta\geq\epsilon$) and outside due to finite $C$.

    \begin{thm}\label{thm:learned_ICCBF_control}
        Consider system \eqref{eq:sys_generic} with input constraints $\U$ and $u_{\rm nom}\in\U$. Let $d$ and $\hat{d}$ be as in \eqref{eq:edf} and \eqref{eq:dhat} and $\sigma>0$ be such that $\max_{z \in Z_{\mathrm{free}}} |\hat d(z)-d(z)|\leq \sigma$. Consider $h$ and $\beta > \sigma$ in \eqref{eq:hat_h}. If $ h_N(x)$ in \eqref{eq:hat_h} is an ICCBF {and $L_g h_{N}(x)$ is full rank} then the control $u_{\mathrm{or}}$ in \eqref{eq:explicit_QP} keeps system~\eqref{eq:sys_generic} in $\Z_{\mathrm{free}}$ {and hence safe}.
        % If $h_N(x)$ is an ICCBF and $L_g \hat h^\beta_{N-1}(\bar x)$ is full rank, then
        % \begin{align}\label{eq:learned_explicit_QP}
        %     u_{\rm or}(\bar x)=\begin{cases} u_{\rm nom}(\bar x)  &\text{if }L_f\hat h^\beta_{N}(\bar x)+L_g \hat h^\beta_{N}(\bar x)u_{\rm nom}\\
        %     & \qquad\qquad  + \alpha_N(\hat h^\beta_{N-1}(\bar x))\geq0,\\
        %     u_{\rm safe}(x)  &\text{otherwise,} \end{cases}
        % \end{align}
        % renders the system safe, where $u_{\rm safe}(x)=-L_g \hat h^\beta_{N}(\bar x)^{-1}(L_f\hat h^\beta_{N}(\bar x)+  \alpha_N(\hat h^\beta_{N}(\bar x))$. If indeed $L_g \hat h^\beta_{N}(\bar x)$ is not full rank the system is still safe if $L_f\hat h^\beta_{N}(\bar x)+ \alpha_N (\hat h^\beta_{N}(\bar x))\geq0$.
    \end{thm}
    \begin{proof}
        {By Lemma \ref{lem:ICCBF_control}, $h(x)\geq0$, so that by \eqref{eq:hat_h} $h(x)=\hat d(Cx)-\beta\geq 0$. Then, by hypothesis and $d(Cx)\geq0$, $d(Cx)\geq \hat d(Cx)-\sigma\implies d(Cx)\geq \beta - \sigma>0$ for all $Cx\in \Z_{\rm free}$}.
    \end{proof}
    {SVMs can approximate any continuous function and finite number of data to any desired accuracy \cite{hammer2003note}. Thus, choosing $\epsilon\leq \sigma$ in \eqref{eq:svm_primal} guarantees $|\hat d(z_i)-d(z_i)|\leq \sigma$ for all data samples, $z_i$.} The assumed bound on $\hat d(z)$ is therefore not a limitation of the proposed learning approach, but rather a statement on the quality of the data.
    %\vspace{-1ex}
    \paragraph*{Calculation of the ICCBF}
    {Theorem \ref{thm:learned_ICCBF_control} builds on the modified model \eqref{eq:sys_aug} with only one input constraint, which is {key} for implementation. With box constraint $|u | \leq u_{\rm max}$ we can directly compute the closed-form solution of the minimization as part of the ICCBF construction in \eqref{eq:hi}, almost everywhere:
    \begin{sizeddisplay}{\small}
    \begin{align*}
        h_i(x) &= \inf_{u\in\U} \{L_fh_{i-1}(x)+L_gh_{i-1}(x)u+\alpha_{i-1}(h_{i-1}(x))\}\nonumber\\
        &=%\begin{multlined}[t]
            L_fh_{i-1}(x)- |L_gh_{i-1}(x)|u_{\rm max}
            +\alpha_{i-1}(h_{i-1}(x)),
        %\end{multlined}    
    \end{align*}
    \end{sizeddisplay}
    and therefore also its required derivatives.} Additionally, the full rank condition of the explicit control law $u_{\rm or}(x)$ simplifies to $L_gh_N(x)\neq 0$, and no matrix inversion is necessary. Next we apply the proposed learned, robust safety filter in simulation and in a real experiment. 

\section{Simulation \& experimental results}\label{sec:sim_exp}

\begin{figure}[t]
            \centering
            \begin{subfigure}[t]{0.425\columnwidth}
                \centering
                % \resizebox{.8\columnwidth}{!}{
                % \includegraphics[width=\columnwidth]{figures/pista_sven-1000.png}}
                \setlength\fheight{.7\columnwidth} 
                \setlength\fwidth{.7\columnwidth}
                \input{figures/image_track.tikz}
                \caption{Image with safe/unsafe region in black/light gray ($300m\times 300m$). \label{fig:image_race_track}}
                % The image shows an area of $100$m$\,\times\, 100$m.
            \end{subfigure}
            \hfill
            \begin{subfigure}[t]{0.5\columnwidth}
                \centering
                \resizebox{\columnwidth}{!}{ \includegraphics[width=\columnwidth]{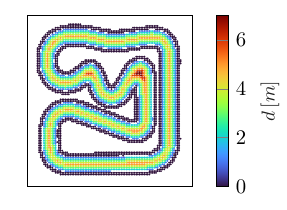}}
                \caption{EDF data $\D$ of $n_s=5444$ training samples with $1m$ resol.}
                \label{fig:edf_data}
            \end{subfigure}
            \hfill
            \caption{Preprocessing for offline learning for closed track we wish to follow. From the image we generate data $\D$.
            %On the left, we have the image which we use to generate samples of the related EDF shown in in the center (Oracle in Figure \ref{fig:diagram}). On the right, we see the zero level sets of these samples as well the learned output function (cf. in Figure \ref{fig:diagram}).
            \label{fig:learning_svm}}
        \end{figure}

\begin{figure}[t]
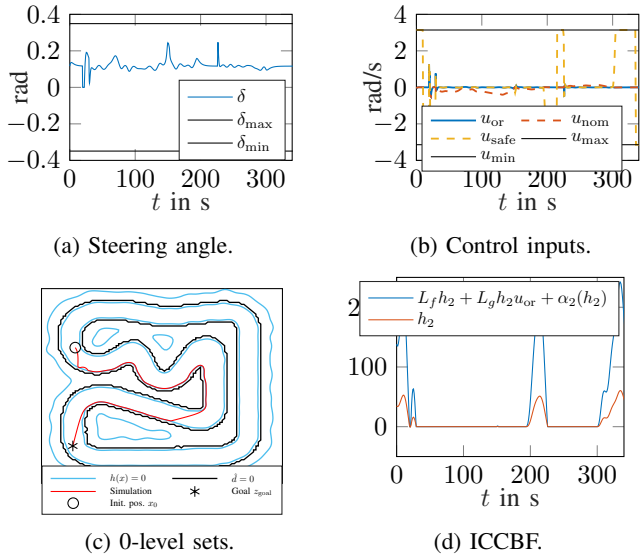

            \centering
            \begin{subfigure}[t]{0.45\columnwidth}
                \centering
                \setlength\fheight{.5\columnwidth} 
                \setlength\fwidth{.8\columnwidth}
                \input{figures/sim_cl_delta.tikz}
                \caption{Steering angle.}
                \label{fig:sim_delta}
            \end{subfigure}
            \hfill
            \begin{subfigure}[t]{0.45\columnwidth}
                \centering
                \setlength\fheight{.5\columnwidth} 
                \setlength\fwidth{.8\columnwidth}
                \input{figures/sim_cl_u.tikz}
                \caption{Control inputs.}
                \label{fig:sim_u}
            \end{subfigure}
            \hfill
            %\vspace{-1cm}
            \begin{subfigure}[t]{0.5\columnwidth}
                \centering
                \setlength\fheight{.6\columnwidth} 
                \setlength\fwidth{.78\columnwidth}
                \input{figures/levelsets_22-06-03}
                \caption{0-level sets. %SVM with $n_{\rm sv}=1737$, score R$^2=0.9823$ and maximum abs. error of $1.01$m. %Hence, we choose robustness parameter $\beta=1.01$.
                }
                \label{fig:learned_cbf}
            \end{subfigure}
            \hfill
            \begin{subfigure}[t]{.46\columnwidth}
                \centering
                \setlength\fheight{.6\columnwidth} 
                \setlength\fwidth{.8\columnwidth}
\definecolor{mycolor1}{rgb}{0.00000,0.44700,0.74100}%
\definecolor{mycolor2}{rgb}{0.85000,0.32500,0.09800}%
\begin{tikzpicture}

\begin{axis}[%
width=0.951\fwidth,
height=\fheight,
at={(0\fwidth,0\fheight)},
scale only axis,
xmin=0,
xmax=340,
xlabel style={font=\color{white!15!black}},
xlabel={$t$ in s},
xlabel shift = -5pt,
ymin=-50,
ymax=250,
ylabel shift=-10pt,
axis background/.style={fill=white},
legend style={legend cell align=left, align=left, draw=white!15!black,nodes={scale=0.65, transform shape}}
]
\addplot [color=mycolor1]
  table[row sep=crcr]{%
0	138.18674661114\\
0.0611510385963925	137.843374411289\\
0.617069571290813	135.524582544275\\
1.61706957129081	135.08164220111\\
2.61706957129081	139.196317728006\\
3.61706957129081	147.185367775153\\
4.61706957129081	158.103688875092\\
7.61706957129081	196.177568418317\\
8.61706957129081	205.774664504412\\
9.61706957129081	211.269504061908\\
10.6170695712908	210.477833718615\\
11.6170695712908	202.395672141843\\
12.6170695712908	187.918823384289\\
13.6170695712908	167.386333347634\\
14.6170695712908	141.710323362908\\
16.6170695712908	80.7676590138279\\
18.8477135771589	12.101116455241\\
19.3259044524394	0\\
20.6824420114827	0\\
21.6824420114827	16.7129102405057\\
22.426704103772	37.4585559346436\\
22.9299767843436	46.981244623659\\
23.4552480300476	54.9380681045027\\
24.2528702283255	62.3655026728169\\
25.2528702283255	63.2448094903151\\
26.2528702283255	55.4091350870184\\
27.2528702283255	40.6094637315\\
28.9012561600961	7.41702153081809\\
29.3166927210765	0\\
149.39129852781	0.0560606677697706\\
150.917794728339	1.1668668473755\\
151.917794728339	0.684506218825504\\
152.203044918103	0.132774766340106\\
152.406790611373	0\\
193.760135015563	0.0200598412717454\\
194.274727151374	0.287022688179547\\
195.274727151374	1.90146414768247\\
196.274727151374	5.56761241275285\\
197.274727151374	11.8043091824728\\
198.274727151374	20.8696100723158\\
199.274727151374	32.7360678865386\\
200.274727151374	47.0856661422094\\
202.274727151374	80.7538863302853\\
204.274727151374	115.802709821639\\
205.274727151374	132.060979640958\\
206.274727151374	146.872100594896\\
207.274727151374	160.051546353963\\
208.274727151374	171.578269857531\\
209.274727151374	181.508829230337\\
210.274727151374	189.882481107816\\
211.274727151374	196.643466990543\\
212.274727151374	201.599840265943\\
213.274727151374	204.426800104931\\
214.274727151374	204.71048552467\\
215.274727151374	200.643577985008\\
216.274727151374	192.426891940111\\
217.274727151374	180.781301987069\\
218.274727151374	165.731078689804\\
219.274727151374	147.518404065122\\
220.274727151374	126.608638894867\\
222.274727151374	79.5348167911689\\
224.274727151374	31.415040693512\\
225.723976611987	0.0604941491552609\\
226.013217051834	0\\
300.147968585584	0.0244952335947346\\
300.640547783754	0.295219203070644\\
301.640547783754	2.07162602255011\\
302.640547783754	6.12510711494792\\
303.640547783754	12.7106554548395\\
304.640547783754	21.4275622066688\\
306.640547783754	41.1047512086635\\
307.640547783754	49.5287923165047\\
308.640547783754	55.6455766963582\\
309.640547783754	59.0607590352948\\
310.640547783754	60.0193698277653\\
311.640547783754	59.3435336460433\\
312.640547783754	58.2380018614778\\
313.640547783754	58.0145142257152\\
314.640547783754	59.8023749150126\\
315.640547783754	64.3124135802268\\
316.640547783754	71.7059234464993\\
317.640547783754	81.5937790645756\\
319.640547783754	105.377810233373\\
320.640547783754	117.25994714254\\
321.640547783754	128.0910213986\\
322.640547783754	137.579504839476\\
324.640547783754	153.607385797585\\
325.640547783754	161.467486560393\\
326.640547783754	170.208939071343\\
327.640547783754	180.295352866215\\
328.640547783754	191.752390751168\\
330.640547783754	216.418142795377\\
331.640547783754	227.49751507121\\
332.640547783754	236.093202458958\\
333.640547783754	241.16445074367\\
334.640547783754	242.075940473086\\
335.640547783754	238.702471040107\\
336.640547783754	230.400483628057\\
337.640547783754	218.699462758172\\
338.640547783754	204.770238459129\\
341	167.924008109459\\
};
\addlegendentry{$L_f h_2+L_g h_2 u_{\rm or}+\alpha_2( h_2)$}

\addplot [color=mycolor2]
  table[row sep=crcr]{%
0	34.9726171724319\\
0.0611510385963925	34.8704267025779\\
0.617069571290813	34.1327117686955\\
1.61706957129081	33.7237177166172\\
2.61706957129081	34.4759653168643\\
3.61706957129081	36.244512076927\\
4.61706957129081	38.8075709088653\\
6.61706957129081	45.1818518146119\\
7.61706957129081	48.3133564353258\\
8.61706957129081	50.8984360246607\\
9.61706957129081	52.5508357287838\\
10.6170695712908	52.8331220971041\\
11.6170695712908	51.2040119513395\\
12.6170695712908	47.983096155025\\
13.6170695712908	43.2177310611576\\
14.6170695712908	37.1012446864597\\
15.6170695712908	29.9521600718779\\
19.388793662095	1.14939773745579\\
19.6751694555957	0.365312448787847\\
19.7152149988882	0.24688254391117\\
20.0706284753154	-0.490173983325803\\
20.3765352433991	-0.664588528316415\\
20.6824420114827	-0.417977622683566\\
21.6824420114827	3.28413339302756\\
22.6132282045606	9.00857769376853\\
23.4552480300476	12.7984192951633\\
24.2528702283255	15.0968413068047\\
25.2528702283255	15.9005952535255\\
26.2528702283255	14.4645917540462\\
27.2528702283255	11.1657336625462\\
28.2528702283255	6.56444926908955\\
29.4497489491172	0.593904062567105\\
29.5828051771579	0.348775699607813\\
29.8285168972508	0.131498121072013\\
30.3551602518573	0.0164896632317095\\
32.8882400115175	8.91752533789258e-05\\
149.917794728339	0.0608023915532954\\
150.917794728339	0.25335251293501\\
151.917794728339	0.233832681070453\\
152.610536304642	0.0276058133466108\\
154.188755203174	-0.000196069647529384\\
194.274727151374	0.0376979815940786\\
195.274727151374	0.346658411241151\\
196.274727151374	1.12525999708788\\
197.274727151374	2.51795915187535\\
198.274727151374	4.60601188788019\\
199.274727151374	7.40066776153344\\
200.274727151374	10.8403551246098\\
201.274727151374	14.7963895880113\\
205.274727151374	32.0197781581938\\
206.274727151374	35.8171991033917\\
207.274727151374	39.2152147098658\\
208.274727151374	42.1995296234102\\
209.274727151374	44.7809383454257\\
210.274727151374	46.9717323018405\\
211.274727151374	48.7648317799104\\
212.274727151374	50.1210030351352\\
213.274727151374	50.9668994896365\\
214.274727151374	51.2036406014506\\
215.274727151374	50.5197062744376\\
216.274727151374	48.6732705935952\\
217.274727151374	45.9771421872281\\
218.274727151374	42.4250609332278\\
219.274727151374	38.0637021355286\\
220.274727151374	32.9956077790674\\
222.274727151374	21.4026134910158\\
224.274727151374	9.31682919420138\\
225.868596831911	0.729712671593575\\
226.013217051834	0.409145536185463\\
226.234390857613	0.169336407966114\\
226.786608630686	0.0189344051204898\\
229.072968736389	-3.06543044530372e-05\\
300.640547783754	0.0383041908691553\\
301.640547783754	0.375250648935207\\
302.640547783754	1.23811994806624\\
303.640547783754	2.72828127020091\\
304.640547783754	4.78620929258324\\
307.640547783754	11.8869793807168\\
308.640547783754	13.5705836200136\\
309.640547783754	14.5935852392178\\
310.640547783754	14.977270220519\\
311.640547783754	14.8933949164003\\
312.640547783754	14.623196440156\\
313.640547783754	14.4925626649958\\
314.640547783754	14.7994378950459\\
315.640547783754	15.7505295725014\\
316.640547783754	17.42139405942\\
317.640547783754	19.747964362264\\
318.640547783754	22.5500677607424\\
320.640547783754	28.5852134298442\\
321.640547783754	31.366274659641\\
322.640547783754	33.8225490354432\\
323.640547783754	35.968823923011\\
326.640547783754	41.9867920740899\\
327.640547783754	44.4207276039533\\
328.640547783754	47.2035164482075\\
331.640547783754	56.2128906402838\\
332.640547783754	58.5353287864423\\
333.640547783754	60.0364406593695\\
334.640547783754	60.5297021343079\\
335.640547783754	59.9513375950337\\
336.640547783754	58.1890581110408\\
337.640547783754	55.4472703172915\\
338.640547783754	52.0892774492663\\
341	42.9803695405858\\
};
\addlegendentry{$ h_2$}

\end{axis}

\begin{axis}[%
width=1.227\fwidth,
height=1.227\fheight,
at={(-0.16\fwidth,-0.135\fheight)},
scale only axis,
xmin=0,
xmax=1,
ymin=0,
ymax=1,
axis line style={draw=none},
ticks=none,
axis x line*=bottom,
axis y line*=left
]
\end{axis}
\end{tikzpicture}%
                \caption{ICCBF.}
                \label{fig:sim_h}
            \end{subfigure}
            \hfill
            \caption{Simulation results. By observing the simulated travelled path in closed loop we notice that the safety filter keeps the robot safe even though nominal control steers directly off track towards the goal $z_{\rm goal}$. The resulting steering angle and rate remain within bounds. The safety filter overrides the nominal control whenever the ICCBF $h_2 (x)$ approaches zero (the robot approaches the road edges), see Figures \ref{fig:sim_u} and \ref{fig:sim_h}. The safety filter is mostly engaged, except in the three periods in which the nominal steering command drives the robot away from the edges (see Figure \ref{fig:learned_cbf}). Importantly, the track itself represents a simply connected set.
            %The related inequality greater than zero suggests that $h_2 (x)$ is a local ICCBF.
            \label{fig:sim}}
            \vspace{-.5cm}
        \end{figure}

    %Towards empirical evidence for the suitability of our proposed method, 
    The simulation and experiment focus on the application of lane-keeping and obstacle avoidance.
    %\vspace{-1ex}
    \paragraph*{Simulation}
        Consider the track shown in Figure \ref{fig:image_race_track} and suppose the car-like robot seeks to move towards some goal $z_{\rm goal}$ without information about the road.
        %, by applying the steering feedback law
        % %\vspace{-.35cm}
        % \begin{align}
        %     u_{\rm nom}=-k_{1}(\delta-\delta_{\rm des}),
        % \end{align}
        % with desired steering $\delta_{\rm des}=-k_2(\theta-\theta_{\rm des})$, where $\theta_{\rm des}=\arctan([z_{\rm goal}-z]_2/[z_{\rm goal}-z]_1)$, and some $k_1,k_2>0$. 
        To avoid moving off road we implement the safety filter discussed above. From the image in Figure \ref{fig:image_race_track} we generate EDF data $\D$ (Figure \ref{fig:edf_data}) 
        % (see e.g. \cite{fabbri20082d}) 
        and learn $\hat d(z)$ offline. We optimize the learning hyperparameters based on a $10$-fold cross-validated grid-search\footnote{The selected hyperparameters are $G=7,\epsilon =.1,\gamma=30$.} using 50-50 split for training and validation, resulting in $n_{\rm sv}=1737$ support vectors and $R^2$ score of 0.9823 with max. abs. error of $1.01$m. 
        %For the dataset being sparse but noise free we set the robustness parameter $\beta$ to the maximum absolute error for the validation set.
        The trajectory obtained with the proposed safety filter successfully remains within the road as shown in Figure \ref{fig:learned_cbf}.
        % The steps are illustrated in Figure~\ref{fig:learning_svm}.
        %
        % Therein, in Figure \ref{fig:learned_cbf} we observe that the maximum error occurs in the sharp turn on the inner side, but prefer these (we can adjust it by $\beta$) over a rugged overfitted barrier.
        %\vspace{-1ex}
        \paragraph*{Implementation}
        Our simulations and experiment suggest that
        %Although there is no formal proof of when $N$ is sufficiently large for $h_N(x)$ satisfying Definition \ref{def:ICCBF} (being a valid ICCBF), simulations indicate that locally 
        taking $N=2$ is sufficient for obtaining an ICCBF\footnote{{Note that $N=1$ does not ensure safety in practice here.}}. The corresponding safety filter therefore requires the evaluation of the first three partial derivatives of $h(x)$, which involve computing sums of terms over all $n_{\rm sv}=1737$ support vectors. By vectorizing all computations we significantly decrease computing time: for the simulation on a laptop\footnote{2.5GHz Dual-Core Intel Core i7, 16GB memory, Intel Iris Plus Graphics} the average computing time for $u_{\rm or}(x)$ decreases from $(1.80 \pm 0.59)\times 10^{-1}$s to $(5.84\pm 6.42)\times 10^{-4}$s; this makes our approach suitable for implementation. {The complexity of the calculation of the gradients and higher-order derivatives scales linearly with the number of vectors in the SVM; even in our Python implementation computing time for necessary gradients for $n_{\rm sv}=1.4\times10^5$ support vectors is at $100$Hz. We expect performance to further improve if implemented in compiled languages such as C++} (see {\url{https://github.com/sb-git-cloud/learning_cbf.git}} for vectorized code).  %\cite{sven_code_github}. 
        \paragraph*{Experimental results}
        The experimental setup is as follows. A modified RC car drives on a closed track with different obstacles (see link above for photos). Its nominal control is set so as to always steer straight. We implement and deploy our learned safety filter with the goal to keep the car on track. The developed ROS package %\cite{sven_code_github}
        is run on a Jetson Nano. %\footnote{NVIDIA Maxwell GPU, Quad-Core Proc., 4GB Mem.}. %Since our algorithm assumes samples of the EDF are available 
        We generate a map in form of an occupancy grid obtained via Hector Slam \cite{hector_slam} and add the race track. The modified occupancy grid map is then used to generate samples of the EDF. For position estimation via LiDAR we use \href{http://wiki.ros.org/amcl}{AMCL}. The experimental results are shown in Figure \ref{fig:experiment}, with a video in the link above. %a video is available on \href{https://www.youtube.com/watch?v=U8eZPTDpHEo}{YouTube}.

        \begin{figure}[t]
            \centering
            \begin{subfigure}[t]{0.95\columnwidth}
                \centering
                \setlength\fheight{.28\columnwidth} 
                \setlength\fwidth{.9\columnwidth}
                \input{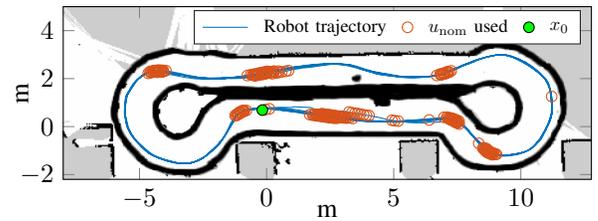}
                \caption{Travelled vehicle path with highlighted $u_{\rm nom}$ when applied.}
                \label{fig:exp_delta}
            \end{subfigure}
            \hfill
            \begin{subfigure}[t]{0.95\columnwidth}
                \centering
                \setlength\fheight{.5\columnwidth} 
                \setlength\fwidth{.9\columnwidth}
                \input{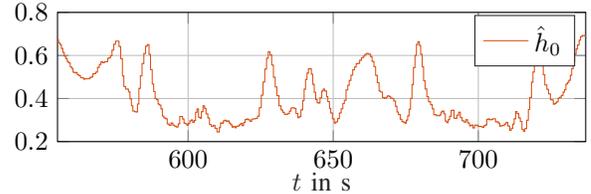}
                \caption{ICCBF for one lap.}
                \label{fig:exp_u}
            \end{subfigure}
            % \hfill
            % \begin{subfigure}[t]{0.23\columnwidth}
            %     \centering
            %     \setlength\fheight{.6\columnwidth} 
            %     \setlength\fwidth{.9\columnwidth}
            %     \input{figures/exp_cl_delta.tikz}
            %     \caption{Steering angle.}
            %     \label{fig:sim_h}
            % \end{subfigure}
            \caption{Experimental results. In Fig. \ref{fig:exp_delta}: the learned safety filter keeps the car on the road while avoiding obstacles and allowing nominal action, $u_{\rm nom}(x)$, (zero steering angle) if safe. Fig. \ref{fig:exp_u} shows that $\hat h_0\geq0$ at all time, {despite input constraints and uncertainties which include: measurement noise from the LiDAR sensor used for map building and localization; limited map resolution (here $5$cm) measurement noise related to the car's velocity; process noise due to discretization of continuous-time model dynamics, and approximated dynamics in general (wheel slip is neglected); input disturbance for the steering angle; limited real-time capabilities of used hardware and ROS.} \label{fig:experiment}}
        \end{figure}

%\vspace{-1ex}
\section{Limitations}
    %SVM does not inherently offer probabilistic interpretations and is problematic for large data sets. The former would be particularly helpful for stochastic analysis and robustness guarantees; though the benefits discussed above particularly for implementation outweigh disadvantages.
    We acknowledge the limitation of a constant velocity; while a varying velocity could generally be accommodated (e.g., by considering acceleration as another input), it would complicate the implementation. A detailed consideration is part of ongoing research. We note that CBF-based controller implicitly assumes that \eqref{eq:QP} always has a feasible solution, i.e., $h_N(x)$ is a valid ICCBF. This is difficult to show in practice, and it is often accommodated with the introduction of slack variables. {Nevertheless safety is still improved compared with standard CBF-based methods.} If safety is not feasible (due to limited steering angle/rate), the explicit control law \eqref{eq:explicit_QP} automatically selects a solution that minimizes the (safety) constraint violation. {Although $\epsilon$-SVM regression for EDFs reduces the risk of rank deficient $L_gh_N(x)$, a theoretical guarantee is generally difficult to obtain.} For simulation and experiment we set the class-$\K$ functions $\alpha_i,i=\{0,1,2\}$, to be positive constants. These constants must be tuned in case of a different vehicle model or different constraints. {This is generally still an open question and interesting directions towards methodologically selecting linear multiples of class-$\K$ functions or extend learning are shown in \cite{9482626,qin-multiagent} for relative-degree one systems.}

\section{Conclusion \& future direction}

We derive a robust filter that guarantees safe operation of car-like robots and present a procedure for its calculation and practical implementation. The proposed design considers practical constraints on the steering angle and the steering rate, and uses environmental data to learn the robust safety filter. Practical use has been illustrated via simulation and implementation on a modified RC car. 
%The related ROS package is provided in \cite{sven_code_github}.
Here the learning was performed offline, but the procedure can be adapted for online or reinforced learning, as well as dynamic environments. Due to space constraints these applications {as well as a complete statistical analysis of all the errors involved in the implementation} are left for future investigations.

%===============================================================================

%===============================================================================

\bibliographystyle{ieeetr}
\bibliography{example}  % .bib

\begin{thebibliography}{10}

\bibitem{aljalbout2021learning}
E.~Aljalbout, J.~Chen, K.~Ritt, M.~Ulmer, and S.~Haddadin, ``{Learning
  Vision-based Reactive Policies for Obstacle Avoidance},'' in {\em Conference
  on Robot Learning}, pp.~2040--2054, PMLR, 2021.

\bibitem{pandey2017mobile}
A.~Pandey, S.~Pandey, and D.~Parhi, ``{Mobile Robot Navigation and Obstacle
  Avoidance Techniques: A Review},'' {\em Int Rob Auto J}, vol.~2, no.~3,
  p.~00022, 2017.

\bibitem{sallab2016end}
A.~E. Sallab, M.~Abdou, E.~Perot, and S.~Yogamani, ``{End-to-End Deep
  Reinforcement Learning for Lane Keeping Assist},'' {\em arXiv preprint
  arXiv:1612.04340}, 2016.

\bibitem{chou2020uncertainty}
G.~Chou, N.~Ozay, and D.~Berenson, ``{Uncertainty-Aware Constraint Learning for
  Adaptive Safe Motion Planning from Demonstrations},'' {\em arXiv preprint
  arXiv:2011.04141}, 2020.

\bibitem{ames:cbf_intro}
A.~D. Ames, S.~Coogan, M.~Egerstedt, G.~Notomista, K.~Sreenath, and P.~Tabuada,
  ``{Control Barrier Functions: Theory and Applications},'' in {\em 2019 18th
  European Control Conference (ECC)}, pp.~3420--3431, 2019.

\bibitem{ctx28136798090006531}
R.~Cheng, M.~J. Khojasteh, A.~D. Ames, and J.~W. Burdick, ``{Safe Multi-Agent
  Interaction through Robust Control Barrier Functions with Learned
  Uncertainties},'' in {\em 2020 59th IEEE Conference on Decision and Control
  (CDC)}, 2020 59th IEEE Conference on Decision and Control (CDC), IEEE,
  2020-12-14.

\bibitem{9303785}
A.~Robey, H.~Hu, L.~Lindemann, H.~Zhang, D.~V. Dimarogonas, S.~Tu, and
  N.~Matni, ``{Learning Control Barrier Functions from Expert
  Demonstrations},'' in {\em 2020 59th IEEE Conference on Decision and Control
  (CDC)}, pp.~3717--3724, 2020.

\bibitem{9294485}
S.~Yaghoubi, G.~Fainekos, and S.~Sankaranarayanan, ``{Training Neural Network
  Controllers Using Control Barrier Functions in the Presence of
  Disturbances},'' in {\em {2020 IEEE 23rd International Conference on
  Intelligent Transportation Systems (ITSC)}}, pp.~1--6, 2020.

\bibitem{lindemann2021learning}
L.~Lindemann, H.~Hu, A.~Robey, H.~Zhang, D.~Dimarogonas, S.~Tu, and N.~Matni,
  ``{Learning Hybrid Control Barrier Functions from Data},'' in {\em Conference
  on Robot Learning}, pp.~1351--1370, PMLR, 2021.

\bibitem{li2021instantaneous}
C.~Li, Z.~Zhang, A.~Nesrin, Q.~Liu, F.~Liu, and M.~Buss, ``{Instantaneous Local
  Control Barrier Function: An Online Learning Approach for Collision
  Avoidance},'' {\em arXiv preprint arXiv:2106.05341}, 2021.

\bibitem{ferlez2020shieldnn}
J.~Ferlez, M.~Elnaggar, Y.~Shoukry, and C.~Fleming, ``{ShieldNN: A Provably
  Safe NN Filter for Unsafe NN Controllers},'' {\em arXiv preprint
  arXiv:2006.09564}, 2020.

\bibitem{barbosa-edf_gauss}
F.~S. Barbosa and J.~Tumuva, ``{Risk-Aware Navigation on Smooth Approximations
  of Euclidean Distance Fields Among Dynamic Obstacles},'' {\em DiVA preprint:
  diva2:1626326}, 2022.

\bibitem{Srinivasan2020SynthesisOC}
M.~Srinivasan, A.~Dabholkar, S.~D. Coogan, and P.~A. Vela, ``{Synthesis of
  Control Barrier Functions Using a Supervised Machine Learning Approach},''
  {\em 2020 IEEE/RSJ International Conference on Intelligent Robots and Systems
  (IROS)}, pp.~7139--7145, 2020.

\bibitem{Oleynikova2016SignedDF}
H.~Oleynikova, A.~Millane, Z.~Taylor, E.~Galceran, J.~I. Nieto, and R.~Y.
  Siegwart, ``{Signed Distance Fields: A Natural Representation for Both
  Mapping and Planning},'' in {\em RSS 2016 Workshop: Geometry and Beyond -
  Representations, Physics, and Scene Understanding for Robotics}, 2016.

\bibitem{taylor:learning_CBF}
A.~Taylor, A.~Singletary, Y.~Yue, and A.~D. Ames, ``{Learning for
  Safety-Critical Control with Control Barrier Functions},'' in {\em {Learning
  for Dynamics and Control}}, pp.~708--717, 2020.

\bibitem{Saveriano:learning_CBF}
M.~Saveriano and D.~Lee, ``{Learning Barrier Functions for Constrained Motion
  Planning with Dynamical Systems},'' in {\em 2019 IEEE/RSJ International
  Conference on Intelligent Robots and Systems (IROS)}, pp.~112--119, 2019.

\bibitem{jin-neural_certificates}
W.~Jin, Z.~Wang, Z.~Yang, and S.~Mou, ``{Neural Certificates for Safe Control
  Policies},'' {\em arXiv preprint: 2006.08465}, 2020.

\bibitem{qin-multiagent}
Z.~Qin, K.~Zhang, Y.~Chen, J.~Chen, and C.~Fan, ``{Learning Safe Multi-Agent
  Control with Decentralized Neural Barrier Certificates},'' {\em arXiv
  preprint: 2101.05436}, 2021.

\bibitem{dawson2021safe}
C.~Dawson, Z.~Qin, S.~Gao, and C.~Fan, ``{Safe Nonlinear Control Using Robust
  Neural Lyapunov-Barrier Functions},'' in {\em {5th Annual Conference on Robot
  Learning }}, 2021.

\bibitem{dawson2022safe}
C.~Dawson, S.~Gao, and C.~Fan, ``{Safe Control with Learned Certificates: A
  Survey of Neural Lyapunov, Barrier, and Contraction methods},'' {\em arXiv
  preprint: 2202.11762}, 2022.

\bibitem{9392327}
K.~Long, C.~Qian, J.~Cortés, and N.~Atanasov, ``{Learning Barrier Functions
  With Memory for Robust Safe Navigation},'' {\em IEEE Robotics and Automation
  Letters}, vol.~6, no.~3, pp.~4931--4938, 2021.

\bibitem{8114339}
X.~Xu, J.~W. Grizzle, P.~Tabuada, and A.~D. Ames, ``{Correctness Guarantees for
  the Composition of Lane Keeping and Adaptive Cruise Control},'' {\em IEEE
  Transactions on Automation Science and Engineering}, vol.~15, no.~3,
  pp.~1216--1229, 2018.

\bibitem{covorsi}
M.~Cavorsi, M.~Khajenejad, R.~Niu, Q.~Shen, and S.~Z. Yong, ``{Tractable
  Compositions of Discrete-Time Control Barrier Functions with Application to
  Lane Keeping and Obstacle Avoidance},'' {\em arXiv preprint: 2004.01858},
  2020.

\bibitem{sven_drew_cbf}
S.~Br{\"u}ggemann, D.~Steeves, and M.~Krstic, ``{Simultaneous Lane-Keeping and
  Obstacle Avoidance by Combining Model Predictive Control and Control Barrier
  Functions},'' {\em arXiv preprint: 2204.06136}, 2022.

\bibitem{8405547}
S.~Kolathaya and A.~D. Ames, ``{Input-to-State Safety With Control Barrier
  Functions},'' {\em IEEE Control Systems Letters}, vol.~3, no.~1,
  pp.~108--113, 2019.

\bibitem{ames2016control}
A.~D. Ames, X.~Xu, J.~W. Grizzle, and P.~Tabuada, ``{Control Barrier Function
  Based Quadratic Programs for Safety Critical Systems},'' {\em IEEE
  Transactions on Automatic Control}, vol.~62, no.~8, pp.~3861--3876, 2016.

\bibitem{ROMDLONY201639}
M.~Z. Romdlony and B.~Jayawardhana, ``{Stabilization with guaranteed safety
  using Control Lyapunov–Barrier Function},'' {\em Automatica}, vol.~66,
  pp.~39--47, 2016.

\bibitem{agrawal2021safe}
D.~Agrawal and D.~Panagou, ``{Safe Control Synthesis via Input Constrained
  Control Barrier Functions},'' {\em arXiv preprint arXiv:2104.01704}, 2021.

\bibitem{Khalil:1173048}
H.~K. Khalil, {\em {{Nonlinear Systems}}}.
\newblock Upper Saddle River, NJ: Prentice-Hall, 3$^\text{rd}$~ed., 2002.

\bibitem{nguyen2016exponential}
Q.~Nguyen and K.~Sreenath, ``{Exponential Control Barrier Functions for
  enforcing high relative-degree safety-critical constraints},'' in {\em {2016
  American Control Conference (ACC)}}, pp.~322--328, IEEE, 2016.

\bibitem{rajamani2011vehicle}
R.~Rajamani, {\em {Vehicle Dynamics and Control}}.
\newblock Mechanical Engineering Series, Springer US, 2011.

\bibitem{xu2018constrained}
X.~Xu, ``{Constrained control of input--output linearizable systems using
  control sharing barrier functions},'' {\em Automatica}, vol.~87,
  pp.~195--201, 2018.

\bibitem{smola2004tutorial}
A.~J. Smola and B.~Sch{\"o}lkopf, ``{A tutorial on support vector
  regression},'' {\em Statistics and computing}, vol.~14, no.~3, pp.~199--222,
  2004.

\bibitem{hammer2003note}
B.~Hammer and K.~Gersmann, ``{A Note on the Universal Approximation Capability
  of Support Vector Machines},'' {\em Neural Processing Letters}, vol.~17,
  no.~1, pp.~43--53, 2003.

\bibitem{hector_slam}
S.~Kohlbrecher, J.~Meyer, O.~von Stryk, and U.~Klingauf, ``{A Flexible and
  Scalable SLAM System with Full 3D Motion Estimation},'' in {\em Proc. IEEE
  International Symposium on Safety, Security and Rescue Robotics (SSRR)},
  IEEE, November 2011.

\bibitem{9482626}
J.~Zeng, B.~Zhang, Z.~Li, and K.~Sreenath, ``{Safety-Critical Control using
  Optimal-decay Control Barrier Function with Guaranteed Point-wise
  Feasibility},'' in {\em 2021 American Control Conference (ACC)},
  pp.~3856--3863, 2021.

\end{thebibliography}

\end{document}